# Interpretable Machine Learning for Self-Service High-Risk Decision-Making

Charles Recaido
*Dept. of Computer Science*
*Cemtral Washington University*
Ellensburg, WA, USA
RecaidoC@cwu.edu

Boris Kovalerchuk
*Dept. of Computer Science*
*Central Washington University*
Ellensburg, WA, USA
BorisK@cwu.edu

*Abstract*—This paper contributes to interpretable machine learning via visual knowledge discovery in general line coordinates (GLC). The concepts of hyperblocks as interpretable dataset units and general line coordinates are combined to create a visual self-service machine learning model. The DSC1 and DSC2 lossless multidimensional coordinate systems are proposed. DSC1 and DSC2 can map multiple dataset attributes to a single two-dimensional (X, Y) Cartesian plane using a graph construction algorithm.  The hyperblock analysis was used to determine visually appealing dataset attribute orders and to reduce line occlusion. It is shown that hyperblocks can generalize decision tree rules and a series of DSC1 or DSC2 plots can visualize a decision tree. The DSC1 and DSC2 plots were tested on benchmark datasets from the UCI ML repository. They allowed for visual classification of data. Additionally, areas of hyperblock impurity were discovered and used to establish dataset splits that highlight the upper estimate of worst-case model accuracy to guide model selection for high-risk decision-making. Major benefits of DSC1 and DSC2 is their highly interpretable nature. They allow domain experts to control or establish new machine learning models through visual pattern discovery.

## I. Introduction

The importance of getting interpretable Machine Learning (ML) models is widely recognized for both traditional and Deep Leaning algorithms and models [1]. While the claims about interpretability of different ML models is an area of active debates [1, 2], the interpretability of Decision Trees (DTs) and logical models, in general, is considered as undisputable.  It implies that DTs and logical models that mimic other ML models offer explanation of those ML models. Moreover, if a DT or a logical model are accurate enough then they can be used in the first place instead of other models with questionable interpretability. These properties of DTs and logical models make them very important for the progress of the interpretable Machine Learning domain.

This paper shows that Visual Knowledge Discovery (VKD) offers a promising approach to analyze, discover and visualize accurate and interpretable DTs and logical models.  It uses multidimensional coordinate systems such as General Line Coordinates (GLC) which offer lossless dimensional reduction, pattern discovery, real-time feature engineering, complex rule creation, and model simplification for domain experts [3].

This paper focuses on the creation and application of two forms of GLC systems referred to as DSC1 and DSC2. The goal of application of these two multidimensional coordinate systems is to give domain experts abilities to control model discovery, analysis, and adjustment. While often domain experts lack technical skills to do this for traditional ML the visual medium makes it feasible allowing domain experts to bring domain expertise and knowledge that is absent in the training data.

Domain experts will be able to control rule creation through VKD using highly interpretable data units known as hyperblocks (HB). This allows domain experts to enhance current models or create new models. The highly interpretable nature of hyperblocks on GLC [4] will provide domain experts with confident decision-making in life-critical or high-risk / high-stakes classification problems.

This paper identifies two aspects of DSC1 and DSC2. The first aspect highlights rule creation through DSC1 and DSC2 by establishing a visual plot series equivalent to a decision tree. The second aspect highlights the VKD process of finding the upper estimate of the worst dataset training-validation split.  The upper estimate of the worst dataset split can influence model selection in high-risk or life-critical problems [3].

Section II presents: (1) plot creation of DSC1 and DSC2 and (2) rule creation on DSC1 and DSC2. We first define the graph construction algorithm (GCA) that generalizes parallel coordinates (PC) for DSC1 and establish classification rules on the Iris dataset [5]. Second, we define the GCA that generalizes shifted paired coordinates (SPC) for DSC2 and establish classification rules of the Iris dataset.

Section III presents training-validation splitting method using DSC2 on the Wisconsin Breast Cancer (WBC) dataset [5]. We are interested in finding the upper estimate of the worst case split to show model reliability of eight different traditional machine learning methods and model selection for high-risk problems. We compared of average model accuracy obtained through tenfold cross validation and worst model accuracy using an upper estimate of the worst-case training-validation split. It had shown the significant difference in accuracy that impacts the best model selection for the high-risk applications.

The approaches in Section II and Section III were implemented on GLC-Vis software using Python as backend language support and OpenGL for visualization rendering. Section IV presents conclusion and future studies into the development of GLC systems and interpretable machine learning.

## II. INTERPRETABLE HYPERBLOCK DATA UNITS ON GENERAL LINE COORDINATES

### A. Hyperblocks as Interpretable Data Units

In the simplest form a hyperblock (HB) is a rectangle in multidimensional space. Hyperblocks are highly interpretable data units as a combination individual dimensions without non-interpretable operations with them. This allows for the separation of different data units (e.g., cell count vs. cell size) during model creation like a decision tree. In a previous study hyperblocks were used to generalize decision tree rules [4].

Definition. A hyperblock is a multidimensional "rectangle" (n-orthotope) with a set of multidimensional points $\{x = (x_1, x_2, ..., x_n)\}$ with center $c = (c_1, c_2, ..., c_n)$ and lengths $L = (L_1, L_2, ... L_n)$ such that,

$$\forall i \in N, |x_i - c_i| \leq \frac{L_i}{2} \qquad (1)$$

Three properties of hyperblocks are pertinent to this paper:
- Hyperblocks are given a purity rating with the highest purity HBs containing samples from only one class.
- Multidimensional datasets can be broken down into a union of non-overlapping pure hyperblocks.
- If attributes $X_1, X_2, ..., X_n$ of n-D space are interpretable then hyperblocks defined in (1) are interpretable.

Methods like principal component analysis can produce HBs but those HBs can be non-interpretable. Non-overlapping interpretable hyperblocks from a specific dataset can be found using decision trees, Merger Hyperblock algorithm [4] and by other methods. For model demonstration we used decision tree based HBs due to time savings as creating a full decision tree is done with linear time complexity.

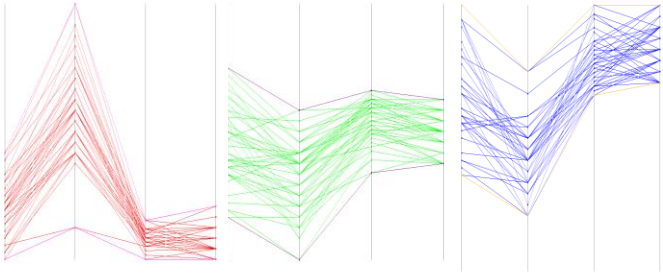

(a) Setosa HB.   (b) Versicolor HB.   (c) Virginica HB.
Fig. 1. Isolated hyperblocks on 4-D PC plot.

Using decision tree analysis shown in Fig. A1 in the appendix, we were able to split the Iris dataset into eight pure non-overlapping hyperblocks. The red shaded block contains all 50 samples of the Setosa class (Fig. 1a), the green shaded block contains 47 samples of the Versicolor class (Fig. 1b), and the blue shaded block contains 43 samples of the Virginica class (Fig. 1c). The six purple shaded blocks contain the remaining 10 samples of the Virginica and Versicolor classes.

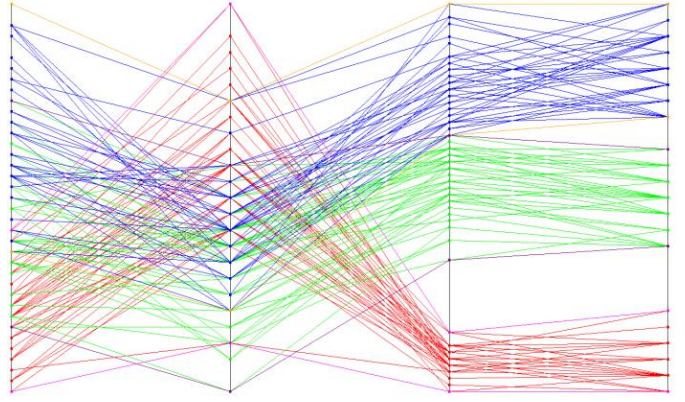

Fig. 2. Three hyperblocks shown together on 4-D PC plot.

Figs. 1 and 2 demonstrate the three largest pure non-overlapping hyperblocks of the Iris dataset on parallel coordinates containing 140 out of 150 samples. The remaining 10 samples were split among six additional hyperblocks and not shown.

The three hyperblocks in Fig. 2 are separated in 4-D space, however, the parallel coordinate plot has a large amount of line overlap within the first and second dimension. To show complete visual separation of non-overlapping hyperblocks we developed DSC1 and DSC2. DSC1 has a limitation in that it can only guarantee complete visual separation of two non-overlapping hyperblocks. However, multiple DSC1 plots can be shown side by side to exhibit other hyperblock pairings. DSC2 can guarantee complete visual separation of three non-overlapping hyperblocks using a non-linear scaling technique.

### B. DSC1 Graph Construction Algorithm

<u>D</u>ynamic <u>S</u>caffold <u>C</u>oordinates using parallel coordinates (DSC1) is a lossless multidimensional coordinate system. DSC1 reduces the *n* axes required in parallel coordinates to a single pair of axes.

The graph constructing algorithm is as follows:

Step 1 to 3 are applied to the entire attribute.
Step 4 to 6 are applied to the individual sample.

(1) Scale each attribute independently within the same range such as [-1, 1] or [0, 1].
(2) Rotate the first attribute by -10°.
   a. x-components = cos(90°-10°)*attribute values
   b. y-components = sin(90°-10°)*attribute values
(3) Rotate the remaining attributes by -45°.
   a. x-components = cos(90°-45°)*attribute values
   b. y-components = sin(90°-45°)*attribute values
(4) Create scaffold lines from the attribute origins to the attribute points.
(5) Connect the first attribute scaffold to the DSC1 plot origin.
(6) Connect the remaining attribute scaffolds from tip-to-tail
(7) Repeat steps 4 to 6 for all samples.
Fig.3 illustrates this process.

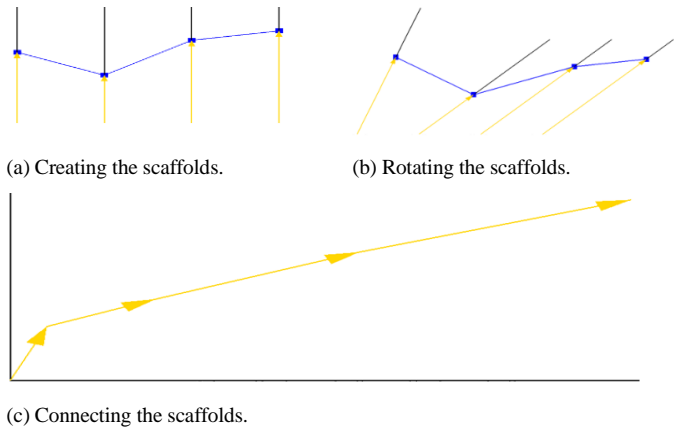

(a) Creating the scaffolds.    (b) Rotating the scaffolds.

(c) Connecting the scaffolds.

Fig. 3. DSC1 graph constructing algorithm simplified for a 4-D point.

The angles in the DSC1 graph construction algorithm are chosen to visually show separation of two non-overlapping hyperblocks. Non-overlapping hyperblocks are guaranteed to be separated on at least one attribute which is referred to as the *attribute of separation*. This attribute of separation is placed first in the order of attributes and given the steepest angle to emphasize its importance. In the case of multiple attributes of separation between any two hyperblocks only one needs to be chosen. The order for the remaining attributes does not matter and share the same angle. In Fig. 3 the scaffold tips are shown to retain all information of the sample. GLC-Viz allows for manipulation of each axis rotation should the user want to develop other DSC1 based models.

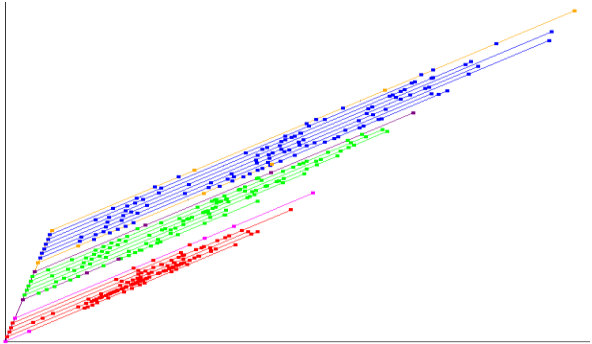

Fig. 4. Three non-overlapping pure HBs from the Iris dataset on DSC1.

Fig. 4 demonstrates the three hyperblocks from Fig. 1 in DSC1. The three hyperblocks are all separated on the fourth attribute line in parallel coordinates as show in Fig. 2. The fourth attribute was given the 1$^{st}$ spot in the attribute order followed by attributes 2, 3, and 1. Only one DSC1 plot is required to demonstrate these three hyperblocks as they share the same attribute of separation. DSC1 is an excellent tool for the Iris dataset as 140 samples can be separated with only the petal width attribute. However, separating the remaining 10 samples (not shown in Fig. 4) requires a DSC1 series as three attributes of separation are required.

DSC1 grants the user the ability to reduce sample lines to an upper and lower hyperblock boundary line as shown in Fig. 5. The alternative visualization reduces line occlusion, which can enhance visual knowledge discovery in sample dense but highly separable datasets. Hyperblocks containing only a few samples relative to the size of dataset do not benefit from this technique.

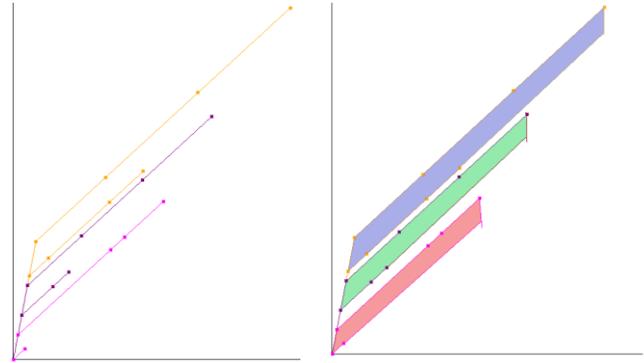

(a) HB boundaries.    (b) Shaded HBs.

Fig. 5. DSC1 visualized using only HBs.

### C. Rule Creation on DSC1

DSC1 plot series can act as a visual aid to highly interpretable models using graphically linear separators. We used the Iris dataset decision tree rules as shown in Fig. A1 in the appendix to establish graphically linear separators on DSC1 plot series.

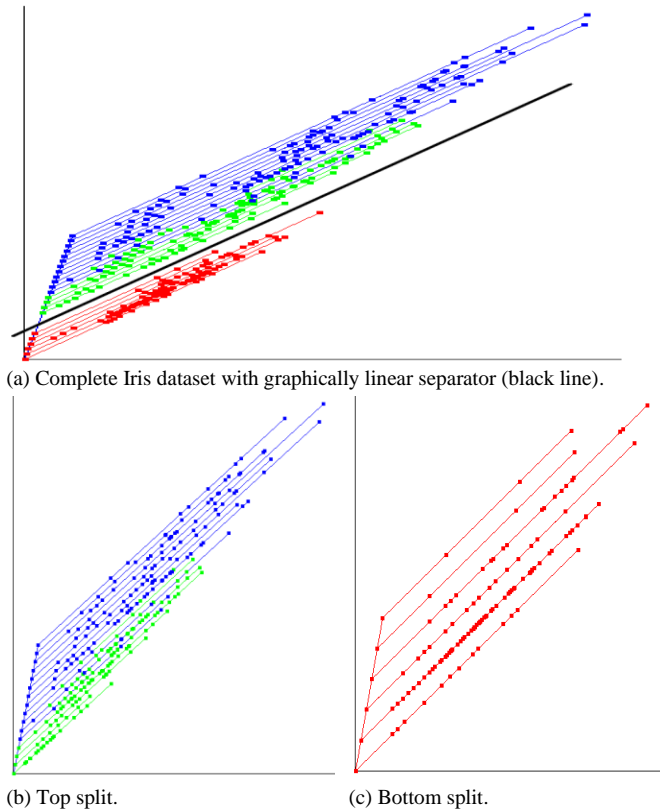

(a) Complete Iris dataset with graphically linear separator (black line).

(b) Top split.    (c) Bottom split.

Fig. 6. Iris dataset after 1$^{st}$ rule is applied.

Fig. 6 shows the graphically linear separator based on the first rule of the decision tree which used the petal width attribute to separate the entire Setosa class from the Virginica and Versicolor classes.

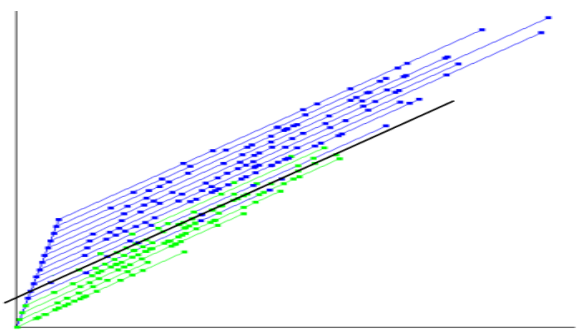

Fig. 7. Iris dataset after 2$^{nd}$ rule is applied.

Fig. 7 shows the graphically linear separator based on the second rule of the decision tree, which used the petal width attribute once more to separate 98% of the versicolor class with 10% of the virginica class from 90% of the virginica class with 2% of the versicolor class. A complete DSC1 plot series of the decision tree in Fig. A1 in the appendix would require eight graphically linear separators and seven plots. The reason behind there being one less plot than the number of graphically linear separators is due to Figs. 6 and 7 being sequential rules that use the same attribute of separation. For simplicity they are separated in this paper but can be combined into a single plot with two graphically linear separators. The remaining rules of the decision tree are not sequential and require their own plot. However, continuing the plot series beyond Figs. 6 and 7 could run the risk of overfitting the data. One technique to reduce overfitting in the decision tree is to limit the depth of the tree which would correspond to less DSC1 plots.

We used DSC1 to visualize the decision tree, however, we can use hyperblocks obtained from other hyperblock algorithms such as Merger-Hyperblock [4] to create similar rules using graphically linear separators that are established through optimization.

### D. Shifted Paired Coordinates

Shifted Paired Coordinates (SPC) represent multidimensional data using several two-dimensional axes. Each pair of axes holds the informational space of two attributes. Unlike parallel coordinates, SPC can show relationships between two attributes. A limitation of the SPC plot is an even number of attributes are required. A dataset with odd number of attributes will require engineering, duplicating, or removing an attribute.

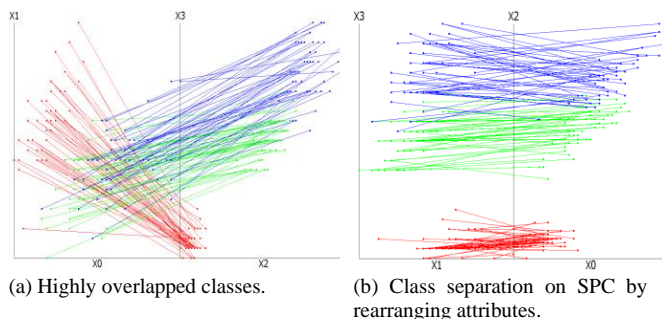

(a) Highly overlapped classes.  (b) Class separation on SPC by rearranging attributes.

Fig. 8. Iris dataset on shifted paired coordinates.

The order of attributes is important when making a SPC plot due to the unique relationship between any two attributes. Fig. 8 shows that a simple rearrangement of attribute-pairs creates a visually appealing plot. This introduces a new hurdle in developing multidimensional visualizations where factorial time complexity would be needed to test every attribute order permutation. Genetic algorithms have been developed to optimize the search space [7]. For this paper we used the decision tree analysis to find hyperblocks separated on attributes of interest to optimize attribute-pair ordering in a similar manner to Section IIB to IIC.

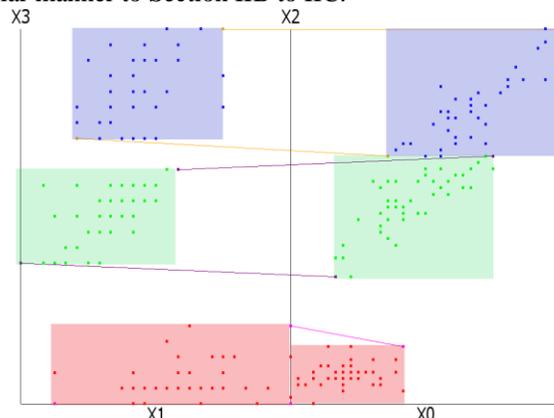

Fig. 9. Iris Hyperblocks on shifted paired coordinates.

Hyperblocks on SPC can be visualized by using several boxes that have their placement dictated by the points on the original maximum-minimum boundary lines. The need for a box arises due to the 2-D basis of the axes-pair. By expanding SPC into a 3-D plot where we use three attribute-pairing a prism would represent the hyperblock boundaries.

### E. DSC2 Graph Construction Algorithm

<u>D</u>ynamic <u>S</u>caffold <u>C</u>oordinates using shifted paired coordinates (DSC2) is a lossless multidimensional coordinate system. DSC2 collapse the n/2 SPC plots to a single pair of axes plot.

The graph constructing algorithm is as follows:

Step 1 is applied to the entire attribute column.
Step 2 to 4 are applied to the individual sample.

(1) Scale each attribute independently within the same range such as [-1, 1] or [0, 1].
(2) Create scaffold lines from the attribute-pair origins to the attribute points.
(3) Connect the first attribute-pair scaffold to the DSC2 plot origin and remove the tail.
(4) Connect the remaining attribute scaffolds from tip-to-tail.
(5) Repeat steps 3-4 for all samples.

Fig. 10 illustrates these steps.

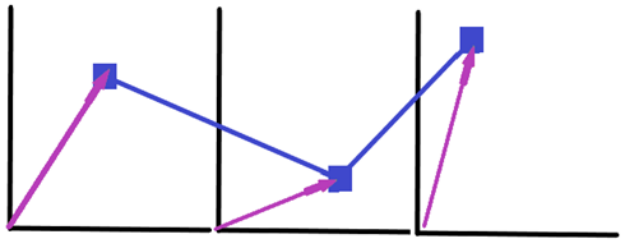

(a) Creating the scaffolds.

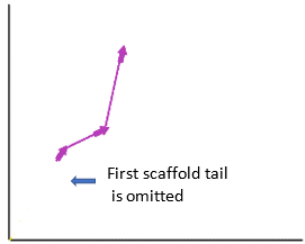

(b) Connecting the scaffolds.

Fig. 10. DSC2 graph constructing algorithm simplified.

When building Fig. 11 we used a decision tree to find two hyperblocks. Their attribute of separation is placed at the beginning of the attribute order. Next, we found a second pairing of hyperblocks that use a different attribute of separation and added their attribute of separation to the second spot in the attribute order. These two attributes make the first attribute-pairing. The remaining attributes can be paired and placed in any order. We want to highlight attributes of interest at the beginning of the plot to influence class separation.

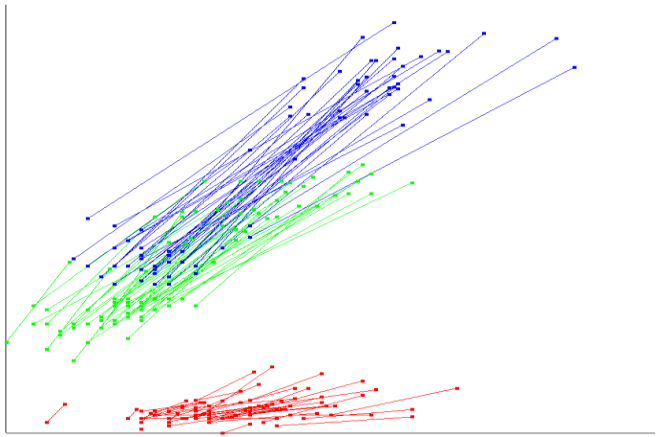

Fig. 11. DSC2 of the Iris dataset.

DSC2 differs from DSC1 in that the hyperblock separation is not immediately seen. The boundary boxes shown in Fig. 9 condense during the scaffolding connection process. Fig. 12 illustrates the boundary boxes from each (X, Y) axes on shifted paired coordinate and how they condense together to cause overlap within the hyperblock space. Hyperblock separation can still be identified by looking at the individual boundary boxes made by each attribute-pairing. The two spots of overlap in Fig. 12 are the first red attribute-pairing overlapped with the second red attribute-pairing and the first blue attribute pairing overlapped with the second green attribute pairing. The first attribute pairing planes are separated for all three hyperblocks and similarly for the second attribute pairing planes.

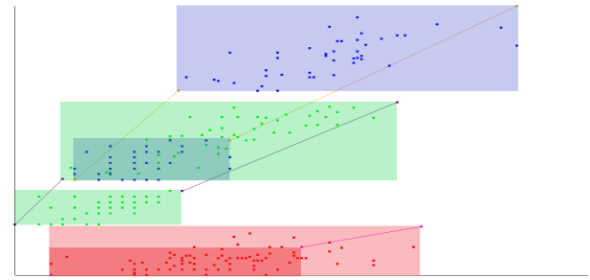

Fig. 12. Hyperblocks on DSC2.

*F. Rule Creation on DSC2*

We developed two techniques to manipulate the DSC2 plot. The first technique is non-linear scaling via graphically linear separators [7] and the second technique is reducing the size of unimportant attributes.

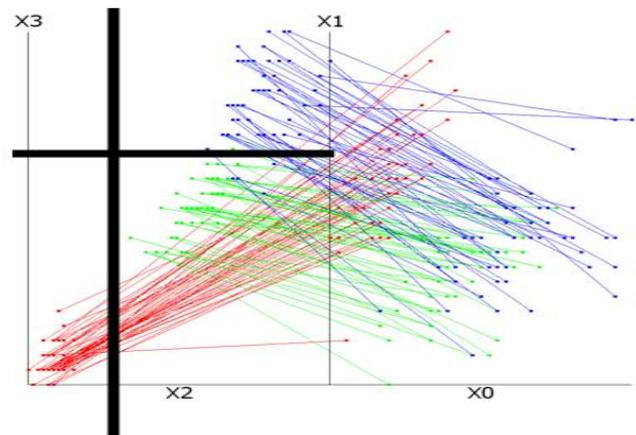

Fig 13. Non-linear scaling technique.

The non-linear scaling technique seen in Fig. 13 is applied before the scaffolding process. The SPC plot allows for both vertical and horizontal linear separators. The points to the left of the vertical separator are shortened, and the points to the right are lengthened. The points above the horizontal separator are lengthened and the points below are shortened. Points in the bottom-left corner will be shortened in both directions, whilst points in the upper-right corner will lengthen in both directions. The placement of graphically linear separators was chosen based off a combination of decision tree rules and plot analysis.

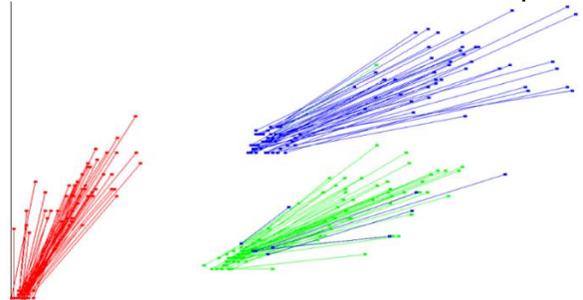

Fig. 14. Iris dataset on DSC2 after non-linear scaling.

Majority of the Iris dataset can be separated on the petal width attribute, but DSC2 allows for two attributes of separation as shown in Fig.14. The vertical graphically linear separator (Fig. 13) was placed at a normalized value of 0.17 by analysis of the DSC1 plot from Fig. 4. On the DSC1 it is shown that the petal length attribute separates the Setosa class. The horizontal graphically linear separator (Fig. 13) was placed at a normalized value of 0.67 by decision tree analysis shown in Fig. A1. The first right node below the root states the decision tree rule where most of the Versicolor class is separated from the Virginica class using petal width attribute.

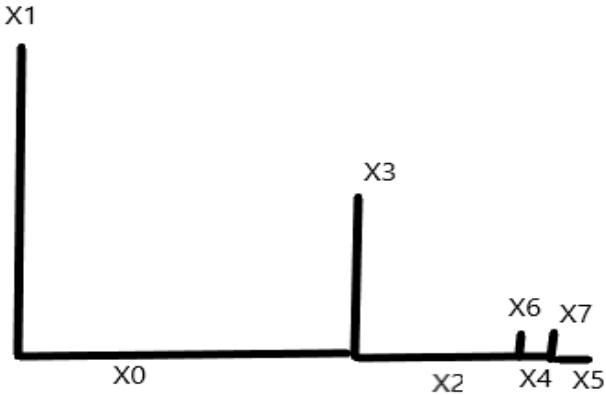

Fig. 15. An example of shrinking attribute-pair axes.

The second technique of reducing size of unimportant attribute does not use graphically linear separators. This technique shrinks the attribute-pair axes of unimportant attributes which results in those attribute-pairs having little influence on the growth of the polylines during the scaffold connection process as show in Fig. 15.

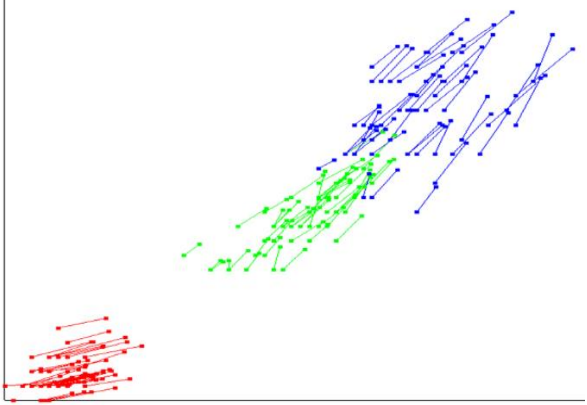

Fig. 16. Downsizing of unimportant attributes on DSC2.

Fig. 16 shows the Iris dataset on DSC2 where the first attribute-pair is given up to 90% of the plot area. The remaining attribute pair is given up to 10% of the plot area. Both techniques can be combined.

### G. Dataset splitting on areas of high overlap

We used DSC1 and DSC2 plot visualization as a visual knowledge discovery tool to find regions of highly overlapped data. These regions exhibit areas that are difficult to separate for classification algorithms. Standard tenfold cross validation relies on random splitting which has no consideration for difficult to classify splits. While random splitting can find a difficult split it is not an effective measure to guarantee a difficult split. Finding a difficult split is crucial in life-critical or high-risk problems that need to know the worst-case performance of a model. In Section III it is shown that some models perform better on a difficult split, but worse on standard tenfold cross validation, which indicates certain models are more beneficial regarding high-risk decisions that may have been previously disregarded dependent on the model selection metric.

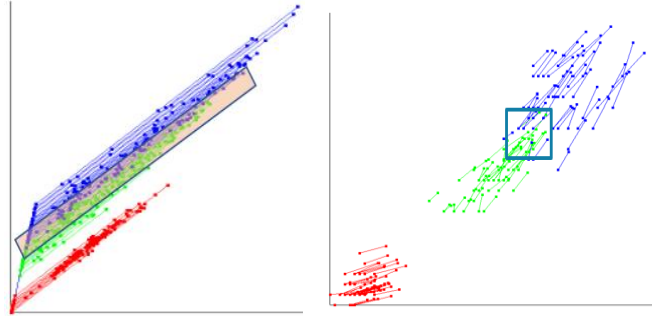

(a) Highly overlapped region on DSC1.    (b) Highly overlapped region on DSC2.

Fig. 17. Finding areas of overlap on DSC based models.

We used a box-bounding (Fig. 17b) or box-clipping algorithm (Fig. 17a) to grab dataset samples from highly overlapped areas and place them into a validation set. The iris dataset only has one area of heavy overlap, but other datasets may require multiple boxes and a decision would need to be made on how many samples to remove from each box.

### III. HIGH-RISK DECISION-MAKING USING WISCONSIN BREAST CANCER DATASET

#### A. Visualizing Wisconsin Breast Cancer Dataset

The Wisconsin Breast Cancer dataset contains 699 samples using nine descriptive attributes [5]. We removed 16 samples which have missing values leaving a total of 683 samples. Those 683 samples include 444 benign cases and 239 malignant cases. We chose the WBC dataset due the high-risk nature of cancer tumor diagnosis. A misdiagnosis of a malignant tumor as a benign tumor could prove fatal for the patient [1].

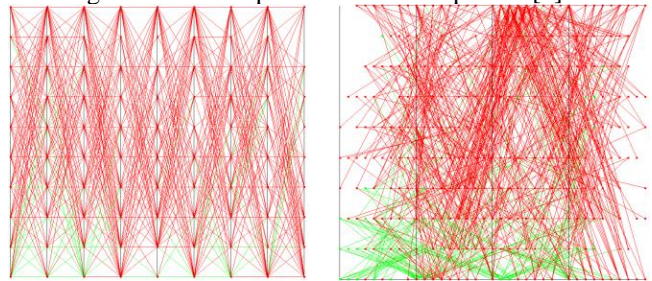

(a) WBC on PC.    (b) WBC on SPC.

Fig. 18. WBC on multidimensional coordinate plots.

Fig. 18 shows that the WBC dataset has heavy overlap when visualized in PC or SPC. One general trend is that the benign cases (green) tend to exist on the bottom of the plot while malignant cases (red) spread over the middle and top region. We guessed that k-Nearest Neighbor (kNN) classifier would have high accuracy on this dataset due to the trend. We used

DSC2 to separate WBC samples by finding two class dominant hyperblocks from a decision tree analysis.

Table I. Hyperblock data from WBC Decision Tree.

|  | Hyperblocks | | | | |
|---|---|---|---|---|---|
|  | HB1 | HB2 | HB4 | HB5 | HB6 |
| Sample Count | 418 | 19 | 34 | 34 | 174 |
| % of Dataset | 61.20 | 2.78 | 4.98 | 4.98 | 25.48 |
| % Purity | 97.13 Benign | 94.74 Benign | 58.82 Malig. | 91.18 Malig. | 98.28 Malig. |
| % of Class | 91.44 | 4.05 | 8.37 | 12.97 | 71.55 |
| Attr. of Influence | $X_1$ | $X_1$-$X_2$-$X_0$ | $X_1$-$X_2$-$X_1$-$X_0$ | $X_1$-$X_2$-$X_1$-$X_0$ | $X_1$-$X_2$-$X_1$ |

Table I shows HB information of the WBC dataset. HB1 and HB6 contain 86.68% of the dataset. We consider these two HBs as the "primary class" HBs. HB1 contains 91.44% of the benign class and HB6 contains 71% of the malignant class. HB1 and HB2 together contain 95.49% of the benign class, and HB3, HB5, and HB6 together contain 86.19% of the malignant class. Attribute $x_1$ was used to separate majority of the benign class. x While both $x_1$ and $x_2$ were used to separate the small number of benign cases that separated with most of the malignant cases which make them excellent attributes of interest.

HB4 and HB3 (not shown in table) are considered low quality HBs. HB4 is a highly impure at 58.82% malignant purity. HB4 could be broken down into smaller and purer HBs, but the DT analysis required more rules and several levels of DT to separate one to two samples at a time. HB3 is very small with only four samples of data. HB3 could be combined with HB2 as they're sibling nodes, however, the parent node has higher impurity. HB3 and HB4 are retained to keep all samples present.

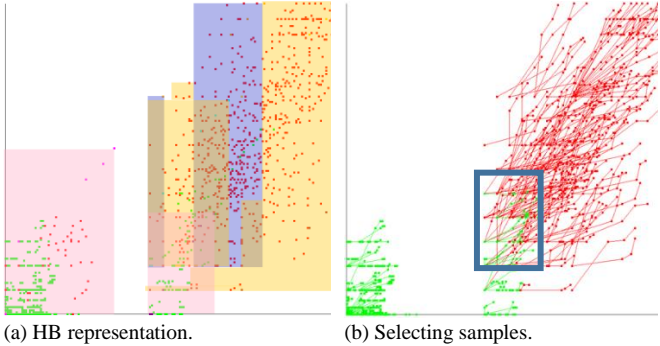

(a) HB representation.  (b) Selecting samples.
Fig. 19. Finding regions of heavy overlap in the WBC dataset.

We placed graphically linear scalars on attributes $x_1$ and $x_2$ as they are attributes of interest on the primary class HBs from Table 1. For the remaining attributes $x_0$, and $x_3$ through $x_6$ we reduced their attribute-pairing to 5% each. The $x_1$ and $x_2$ attribute pairing were expanded by 150% to enhance their influence on class separation. Fig. 19a shows the hyperblock boundaries whilst Fig. 19b shows the WBC dataset with HB impurities removed. A box-boundary algorithm was used to capture 68 samples in an area of heavy overlap to put into a validation set for later testing.

### B. Upper Estimate of the Worst-case Validation Split on Wisconsin Breast Cancer Dataset

Classification results were obtained from eight standard ML classifiers in the sci-kit learn Python library using 10-fold cross validation [8]. The WBC dataset was split into training and validation using the box-bounding area in Fig. 17. The samples in the bounded box were added into the validation set (stopping at 10% of the entire dataset). There were enough overlapped samples to fill the validation set, however, larger validation splits may require drawing samples from non-overlapped areas.

Table II. 10-Fold Cross validation and worst class Results on 8 ML models.

| Model | 10-Fold Cross Validation Accuracy (%) | | | Worst-Split |
|---|---|---|---|---|
|  | Average | Max | Min | Acc (%) |
| DT | 94.7 | 98.5 | 91.3 | 82.5 |
| SVM | 97.1 | 100 | 92.8 | 80.8 |
| RF | 96.6 | 98.5 | 92.8 | 80.8 |
| KNN | **97.2** | 100 | 91.3 | 82.3 |
| LR | 96.8 | 100 | 92.7 | 79.4 |
| NB | 96.1 | 98.5 | 92.7 | **86.7** |
| SGD | 96.2 | 100 | 91.2 | 79.4 |
| MLP | 95.5 | 100 | 89.8 | 79.4 |

Table II shows that standard ML algorithms can classify the WBC dataset within 94 to 97% accuracy without any additional processing, dimensional reduction, or feature engineering. The lowest split was 89.8% and the highest split was 100%. Despite the strong model accuracies obtained in Table II, all eight classifiers had an accuracy between 79% and 86% in the upper estimate of the worst-case scenario. In life-critical and other high-risk applications knowing the worst performance of a model can influence reliance on the model and the possibilities of incorporating additional models into decision-making as a safeguard. In this case 10-fold cross validation accuracy suggests using KNN or support vector machine (SVM) classifier, but the model that performed the best on the estimate of the most difficult split was Naïve Bayes (NB).

## IV. CONCLUSION AND FUTURE WORK

This paper contributes to visual and interpretable machine learning methods by developing DSC1 and DSC2 methods that can be used for multidimensional visualization, analysis, and classification. DSC1 and DSC2 have self-service components that allow domain experts to change, add, or remove attributes and select regions of highly condensed samples for model selection.

Hyperblocks as interpretable data units were used to highlight attributes of separation within a dataset as a computationally efficient alternative to genetic or brute force algorithms for attribute order permutation selection.

The future work will include decreasing the number of visually separate DSC plots that are needed for visualizing the ML classification model rules. It will lift a limitation of the DSC2 algorithm which can guarantee only the separation of three non-overlapping hyperblocks visually. It will be done by connecting multiple plots together using an n-Gon shaped set of axes or other methods such as circle partitioning from the center.

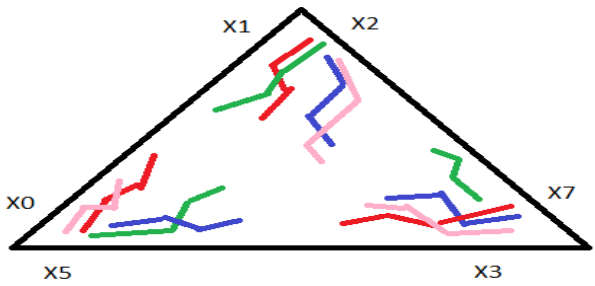

Fig. 20. n-Gon DSC2 visualizing eight dimensions with six attributes of interest and four classes.

Fig. 20 illustrates this idea by showing a combination of three distinct DSC2 plots, which guarantees the separation of nine non-overlapping hyperblocks at the cost of more polylines. In Fig. 20 the origin of each DSC2 plot is at the vertices of the n-Gon (triangle in this example). Each of them visualizes HBs with specific attribute pairs that separates some of them from other HBs. While it increases the polyline count by a factor of the number of vertices in the polygon it allows the separation of multiple attributes within the same plot canvas.

An application for this future focus area would be data like the MNIST handwritten digits dataset. Digit pairs can be separated on a few attributes with high accuracy, but there are 45 different digit pairings to visualize. n-Gon DSC2 visualization will enhance Visual Knowledge Discovery for multi-class problems by expanding the number of attributes of separation that can be represented.

Major benefits of hyperblocks on the general line coordinates in DSC1 and DSC2 is that they enhance the abilities of the end-user and allow them to apply their domain expertise to the model through visual knowledge discovery.

APPENDIX

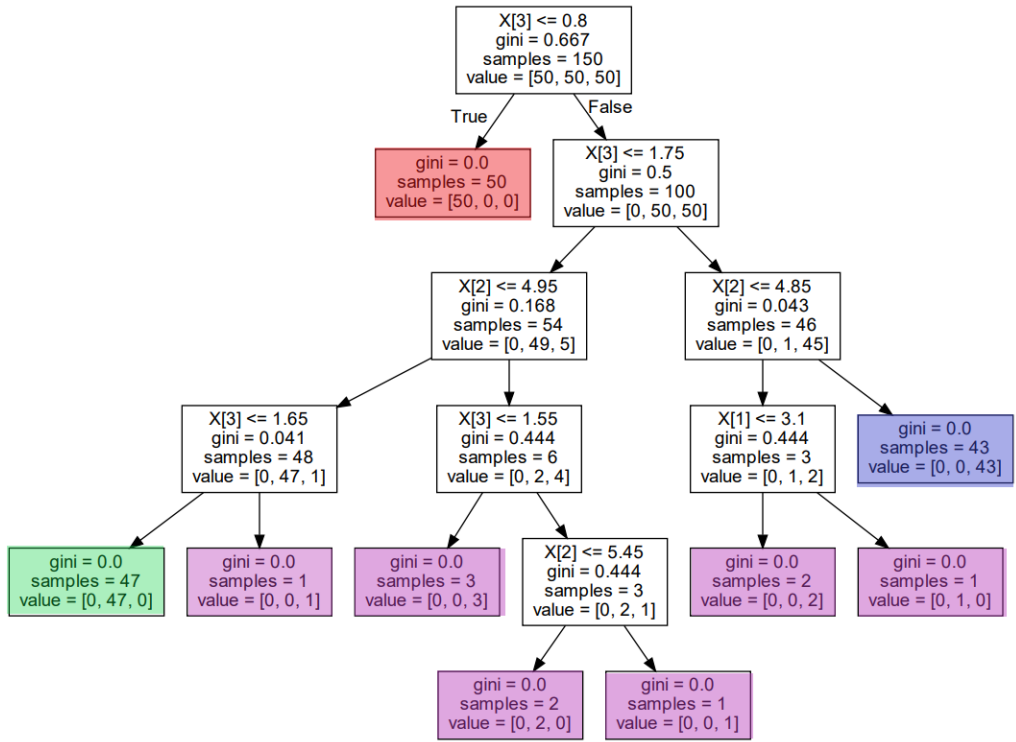

Fig. A1. Iris Dataset Decision Tree.